\documentclass[10pt,twocolumn,letterpaper]{article}

\usepackage[pagenumbers]{cvpr} 

\usepackage{lipsum}
\usepackage{mwe}
\usepackage{graphicx}
\usepackage{multirow}
\usepackage[accsupp]{axessibility}  

\newcommand{\methodname}{ArtLLM\xspace}

\definecolor{cvprblue}{rgb}{0.21,0.49,0.74}
\usepackage[pagebackref,breaklinks,colorlinks,allcolors=cvprblue]{hyperref}

\def\paperID{9316} 
\def\confName{CVPR}
\def\confYear{2026}

\title{\methodname: Generating Articulated Assets via 3D LLM}

\author{
    Penghao Wang$^{1,2,*}$ \quad Siyuan Xie$^{1}$ \quad Hongyu Yan$^{2,3}$ \quad Xianghui Yang$^{2}$ \\ Jingwei Huang$^{2,\dagger}$ \quad Chunchao Guo$^{2,\dagger}$ \quad Jiayuan Gu$^{1,\dagger}$ \\
	$^1${ShanghaiTech University} \quad $^2${Tencent Hunyuan} \quad $^3${HKUST} \\
    \url{https://authoritywang.github.io/artllm}
}

\begin{document}

\twocolumn[{
    \maketitle
    \vspace{-1.0cm}
    \begin{center}
        \includegraphics[width=\linewidth]{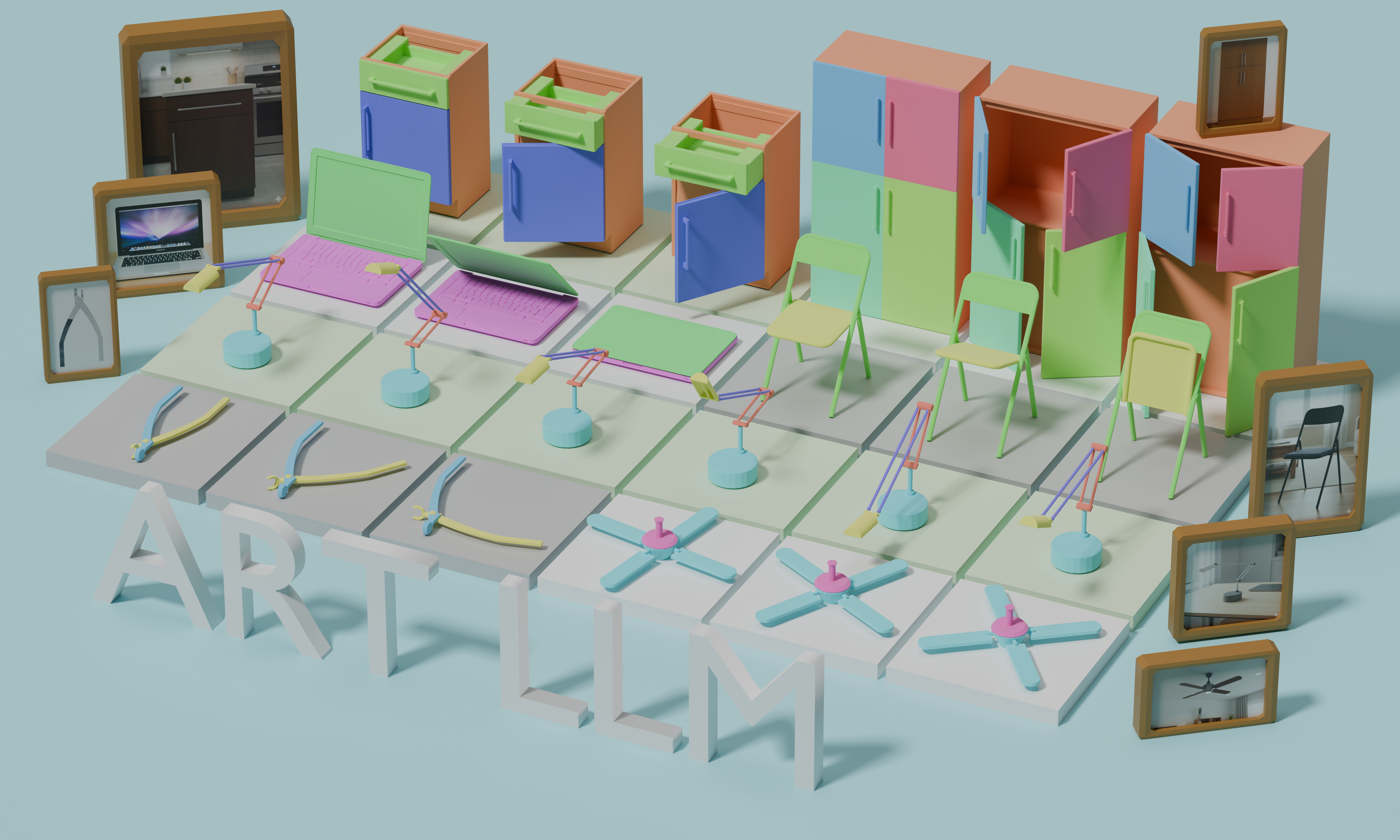}
    \end{center}
    \vspace{-0.5cm}
    \captionsetup{type=figure}
    \captionof{figure}{
        We propose \textbf{\methodname}, a novel framework capable of rapidly generating articulation assets from images or text. By using a 3D LLM to jointly predict part layouts and joints, and integrating state-of-the-art part generation methods, our approach can produce high-quality, physically grounded articulation assets.
    }
    \label{fig:teaser}
    \vspace{0.3cm}
}]

{
    \renewcommand{\thefootnote}{\fnsymbol{footnote}}
    \footnotetext[1]{This work is done while interning with Tencent Hunyuan.}
    \footnotetext[2]{Corresponding authors.}
}


\begin{abstract}

Creating interactive digital environments for gaming, robotics, and simulation relies on articulated 3D objects whose functionality emerges from their part geometry and kinematic structure. However, existing approaches remain fundamentally limited: optimization-based reconstruction methods require slow, per-object joint fitting and typically handle only simple, single-joint objects, while retrieval-based methods assemble parts from a fixed library, leading to repetitive geometry and poor generalization. 
To address these challenges, we introduce \methodname, a novel framework for generating high-quality articulated assets directly from complete 3D meshes. At its core is a 3D multimodal large language model trained on a large-scale articulation dataset curated from both existing articulation datasets and procedurally generated objects. Unlike prior work, \methodname autoregressively predicts a variable number of parts and joints, inferring their kinematic structure in a unified manner from the object's point cloud. This articulation-aware layout then conditions a 3D generative model to synthesize high-fidelity part geometries. Experiments on the PartNet-Mobility dataset show that \methodname significantly outperforms state-of-the-art methods in both part layout accuracy and joint prediction, while generalizing robustly to real-world objects. Finally, we demonstrate its utility in constructing digital twins, highlighting its potential for scalable robot learning. 
\end{abstract}
\section{Introduction}
\label{sec:intro}


The creation of interactive digital worlds for gaming, robotics, and simulation fundamentally relies on assets that can be manipulated and animated. These articulated objects, from doors and drawers to complex machinery, derive their functionality from their underlying part-based geometry and kinematic structures. Generating articulated assets automatically is essential for scaling up content creation, enabling realistic robot training in simulation~\cite{VillasevilSLC0024}, and enriching the interactivity of virtual environments.

Recent efforts in articulated object generation have primarily followed two distinct paradigms. One line of work~\cite{liu2023paris, liu2025videoartgs, liu2025artgs, peng2025generalizable} focuses on optimization-based reconstruction from multi-view images or videos, leveraging neural representations like NeRF~\cite{mildenhall2021nerf} or 3DGS~\cite{kerbl20233d} to estimate joint parameters and geometry. However, these approaches are often hampered by slow, per-object optimization, tend to produce low-fidelity geometry, and are typically constrained to simple objects with only a single joint. Other approaches~\cite{liu2024singapo, liu2024cage} train feedforward networks on existing datasets to directly predict part layouts and joint parameters. While offering fast inference, these methods are usually constrained to retrieving parts from a fixed, predefined database, which severely limits their ability to produce novel shapes and results in geometrically repetitive assets.

While generating high-quality articulated objects remains difficult, general 3D object generation~\cite{zhang20233dshape2vecset,xiang2025structured} has seen immense progress, enabling high-fidelity synthesis from various inputs. Recent extensions support part-level generation~\cite{yan2025x,yang2025omnipart, lin2025partcrafter,tang2025efficient}. Yet, a fundamental limitation persists: a disconnect between geometry and motion. These models are unaware of the underlying kinematic structures that dictate how parts should move, leading to a potential mismatch between a part's visual semantics and its intended mechanical role. It clearly indicates that a unified approach, capable of jointly reasoning over geometry and articulation, is required.

To this end, we present a framework that generates articulated assets by first predicting their geometric layouts as well as kinematic structures and then synthesizing their geometry. The centerpiece is a 3D articulation language model (\textbf{\methodname}), which autoregressively outputs a tokenized blueprint of the object's part layout and kinematic relationships, given an input point cloud. This blueprint then guides a part-aware generative model to synthesize high-fidelity geometries, overcoming the reliance on fixed databases. Furthermore, we introduce a post-processing step to optimize the predicted joint limit, so that the resulting articulation is physically plausible and collision-free. Trained on a large, curated dataset of articulated objects, \methodname offers a scalable and effective solution that avoids the slow optimization of reconstruction methods and the geometric limitations of retrieval-based approaches.

We evaluate our method on the PartNet-Mobility~\cite{xiang2020sapien} dataset and compare it with state-of-the-art approaches. Our model achieves superior performance in predicting part placement, joint accuracy, and kinematic relationship modeling. Unlike retrieval-based methods, it generates accurate and novel part geometries and generalizes well to real-world images, effectively reconstructing articulated assets and enabling realistic digital twins, which highlight its potential to bridge perception and generation for scalable robot learning.


\section{Related Work}%
\label{sec:rw}


\noindent\textbf{3D Generation} 
Early approaches leveraged 2D foundation models for 3D generation, including techniques based on SDS optimization~\cite{poole2022dreamfusion} and multi-view image synthesis~\cite{liu2024one, long2024wonder3d, shi2023mvdream}. Later,  LRM series~\cite{hong2023lrm, tang2024lgm, zhang2024gs, xu2024grm}, introduced a feed-forward paradigm to achieve fast 3d generation. However, their representation is not inherently 3D, which constrained the generation fidelity.
More recently, 3DShape2VecSet~\cite{zhang20233dshape2vecset} and Trellis~\cite{xiang2025structured} respectively introduced native 3D generation representations based on point and voxel, laying the foundation for native 3D generative models. Several works such as ~\cite{zhang2024clay, li2024craftsman3d, li2025triposg, lai2025hunyuan3d, li2025step1x} have further advanced high-quality 3D generation with DiT~\cite{peebles2023scalable}. Building upon these 3D foundation models, many studies have explored downstream 3D generation tasks, including part-level generation~\cite{yan2025x, yang2025omnipart, zhang2025bang, ding2025fullpart, yang2025holopart}, scene-level generation~\cite{huang2025midi, yao2025cast}, diverse condition control generation~\cite{hunyuan3d2025hunyuan3d, yan2025posemaster}, editable generation~\cite{ye2025nano3d}. Collectively, these advancements enable flexible and expressive 3D generation across diverse conditions, establishing robust 3D foundation models that benefit multiple application domains.


\noindent\textbf{3D Large Language Models} Inspired by the success of VLMs~\cite{liu2023visual, bai2025qwen2}, enabling LLMs to understand 3D content has become an urgent and important research direction. Recent works~\cite{hong20233d, xu2024pointllm, qi2024shapellm} have pioneered the direct integration of 3D representations into language models, enabling native 3D reasoning. 
Beyond conversational understanding, subsequent studies have broadened the capabilities of 3D LLMs to tasks such as 3D generation~\cite{wang2024llama, ye2025shapellm, wang2025part}, scene grounding~\cite{mao2025spatiallm}. Collectively, these works demonstrate the practical value and strong potential of 3D LLMs in advancing multimodal understanding and generation tasks.

\begin{figure*}[t]
    \centering
        \includegraphics[width=\textwidth]{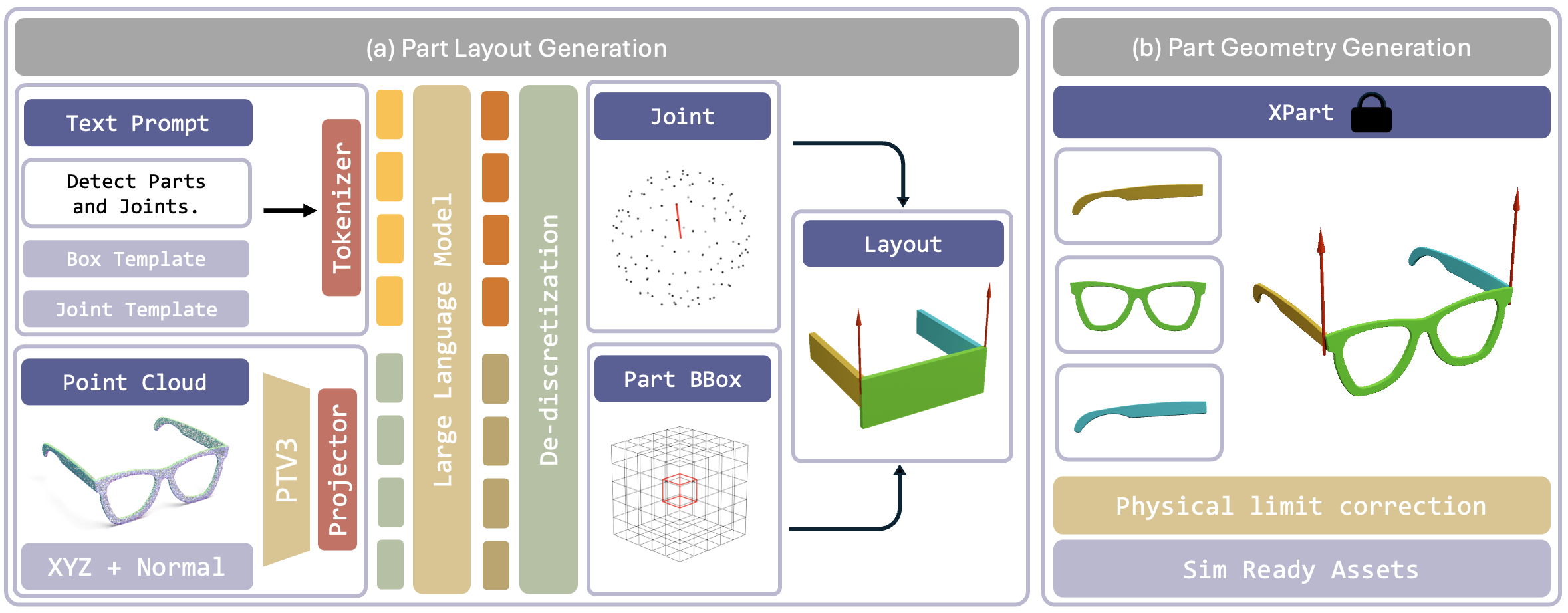}
    \caption{\textbf{Architecture Overview.} Given an input point cloud, \methodname first predicts a tokenized articulation blueprint that specifies \textbf{part layouts and kinematic structures}. This blueprint then conditions a part-aware generative model to synthesize high-fidelity \textbf{link geometries}, followed by a physics-based joint-limit correction module refines the articulation, producing simulation-ready articulated assets. }
    \label{fig:pipeline}
    \vspace{-1.5em}
\end{figure*}

\noindent\textbf{Articulation Assets Generation} Rapidly generating articulated object assets from images or text is crucial for building digital twins and advancing robotic simulation.
Early efforts~\cite{xiang2020sapien, iliash2024s2o, cao2025physx, geng2023gapartnet, mo2019partnet, wang2025partnext, hunyuan3d2026hy3d} relied on manual annotation to construct large-scale part-level and articulated object datasets, laying the foundation for this field.
Recent methods~\cite{liu2023paris, liu2025videoartgs, liu2025artgs, peng2025generalizable} employ per-object optimization to reconstruct articulated objects. However, these approaches suffer from low optimization speed, dependence on dense multi-view or video inputs, and limited scalability.
Other approaches~\cite{liu2024cage, chen2024urdformer, liu2024singapo, le2024articulate, lian2025infinite, wu2025dipo} generate articulated objects by retrieving existing parts from pre-built libraries, which restricts geometric novelty.
To enable novel geometry generation, some studies~\cite{lei2023nap, su2025artformer} perform surface reconstruction from generated SDFs, but the results often lack quality.
With the emergence of 3D foundation models, methods such as ~\cite{lu2025dreamart, chen2025freeart3d, chen2025artilatent} have achieved articulated object generation, though they are typically limited to single-joint structures.
Additionally, some works~\cite{li2025urdf, mandi2024real2code} leverage the strong reasoning ability of LLMs for articulated object modeling. Yet, they directly predict float parameters and are trained on limited data, leading to poor generalization. Their reliance on point cloud based mesh reconstruction further constrains output quality.
In contrast, our approach trains a 3D LLM with a well-designed template and data quantization strategy, enabling accurate prediction of part layouts and joint parameters. Combined with state-of-the-art part generation models, our method produces high-quality articulated objects with high-fidelity geometric structures.
\section{Method}%
\label{sec:method}

This section introduces our framework for generating articulated assets from point clouds. First, our novel 3D articulation language model (\textbf{\methodname}) autoregressively predicts the object's kinematic structure, outputting a tokenized representation (Sec.\ref{sec:ArtLLM}). This model is trained on a large-scale, diverse collection of articulated data (Sec.\ref{sec:ArtLLMDataset}). Next, the previously predicted structural blueprint conditions a part-aware generative model to synthesize high-fidelity, coherent part geometries (Sec.\ref{sec:part_gen}). Finally, a joint-range optimization step ensures the resulting asset is physically plausible and collision-free (Sec.~\ref{sec:limit_correction}). See Figure~\ref{fig:pipeline} for the overall pipeline.

\subsection{3D Articulation Language Model (\methodname)}
\label{sec:ArtLLM}

Kinematic structures of articulated objects are often specified in the \emph{Unified Robotics Description Format} (URDF), which utilizes an XML schema. Thus, we reformulate 3D articulation understanding as a language modeling problem to leverage the powerful reasoning and sequence modeling capabilities of Large Language Models (LLMs). We represent an object's entire kinematic structure—including its constituent parts, their layout, and joint parameters—as a unified sequence of discrete tokens. The autoregressive approach naturally accommodates objects with varying numbers and types of parts, while also allowing us to leverage the rich semantic and structural priors learned by the LLM. Furthermore, this token-based representation is inherently flexible and easily extensible.

\noindent\textbf{Input Representation}
Our model operates on a point cloud representation, allowing it to flexibly handle various input modalities. For text or image inputs, we first leverage off-the-shelf generative models (e.g., Hunyuan3D 2.5~\cite{lai2025hunyuan3d}, TripoSG~\cite{li2025triposg}) to produce an initial 3D mesh. For mesh inputs, whether generated or provided directly, we uniformly sample 32,768 surface points with their corresponding normals. To ensure the consistency of normals, we pre-process the mesh to be watertight before sampling.

\noindent\textbf{Bridging 3D Geometry and Language} To bridge the modality gap between the input point cloud and our LLM, we employ an encoder-projector architecture inspired by the success of vision-language models~\cite{liu2023visual,bai2025qwen2} as well as 3D-language modeling~\cite{qi2024shapellm, xu2024pointllm, mao2025spatiallm}.
We choose Point Transformer v3~\cite{wu2024point} as our point cloud encoder due to its powerful yet efficient design. Following SpatialLM~\cite{mao2025spatiallm}, we augment the final layer's features with position embeddings to preserve crucial spatial information. These augmented features are then projected by a simple two-layer MLP for modality alignment. The Qwen3~\cite{yang2025qwen3} 0.6B model is used as our language model backbone.
\noindent\textbf{Generating Articulation as Languages} We formulate the articulated structure of an object as a language sequence composed of part and joint definitions. We design a concise yet informative text template to regularize outputs. 

Each part is assigned with an \emph{id} and parameterized by its 3D axis-aligned bounding box (AABB):
\begin{equation}
    \scalebox{0.9}{$
    \text{bbox\_}id = \text{BBox}(x_{min},y_{min},z_{min},x_{max},y_{max},z_{max}),
    $}
\end{equation}

Similarly, kinematic joints are defined using a structured format that encodes their type, connectivity, and parameters. We support four primitive types, including \emph{Revolute}, \emph{Continuous}, \emph{Prismatic}, and \emph{Screw}. For example, a revolute joint is written as:
\begin{equation}
    \scalebox{0.9}{$
        \text{joint\_}id = Revolute\text{Joint}(parent, child, dir, pos, limit),
    $}
\end{equation}
where $parent$ and $child$ link are integer IDs of the connected parts; $dir$ and $pos$ are 3D vectors defining the joint's rotation axis and origin; and $limit$ is a tuple specifying the motion range.

The model is designed to autoregressively generate the full sequence, by first predicting all part bounding boxes, followed by a separator token, and then all joint definitions. This ordering ensures that joint prediction is conditioned on the complete part layout, improving structural coherence.

\noindent\textbf{Quantized Predictions} While structured templates simplify the generation task, LLMs are fundamentally designed to predict tokens from a discrete vocabulary, making direct regression of continuous values prone to numerical instability. To address this, we convert all continuous geometric and kinematic parameters into a discrete, token-based representation through quantization. This approach allows us to frame the entire articulation prediction problem within a robust language modeling paradigm.

For each part’s bounding box, we quantize its coordinate from a normalized range of $[-1, 1]$ into discrete 128 bins per axis:
\begin{equation}
\scalebox{0.9}{$
    \hat{c}_{\text{min}} = \left\lfloor \frac{(c_{\text{min}} + 1)}{2} \times 128 \right\rfloor,
    \hat{c}_{\text{max}} = \left\lceil \frac{(c_{\text{max}} + 1)}{2} \times 128 \right\rceil,
$}
\end{equation}
where $c$ represents a original continuous coordinate value and $\hat{c}$ is its its corresponding quantized bin index.



We apply a similar discretization to joint parameters. The joint origin is quantized into 128 bins per axis, identical to the bounding box coordinates. For joint limits, we discretize rotational angles into 48 bins over $[-2\pi, 2\pi]$ and translational distances into 64 bins over $[-2,2]$.
For the joint axis, we create a discrete 128-entry codebook. Motivated by the observation that most joint axes align with the coordinate axes, our codebook is constructed hierarchically. We first sample points uniformly from the unit circles on the XY, YZ, and XZ planes. Then, additional points are obtained via Farthest Point Sampling (FPS) on a Fibonacci sphere. Each point on the unit sphere corresponds to a rotation axis. This codebook design provides dense coverage for axis-aligned directions while maintaining the flexibility to represent other orientations. 
\noindent\textbf{Multi-Task and Multi-Stage SFT} To effectively train our model, we draw inspiration from multi-task learning~\cite{radford2019language}, which improves performance by leveraging shared representations with related auxiliary tasks. The articulation prediction naturally decomposes into two core sub-problems: identifying the geometric layout of its parts and inferring the kinematic relationships (joints) between them. We define three supervised fine-tuning (SFT) tasks:
\begin{enumerate}
\item \textbf{Part Layout Prediction:} Predicts only the part bounding boxes from the point cloud.
\item \textbf{Kinematic Prediction:} Predicts the joints, conditioned on both the point cloud and the ground-truth part layout.
\item \textbf{End-to-End Articulation Prediction:} Predicts both parts and joints from the input point cloud.
\end{enumerate}

Furthermore, we propose a progressive, two-stage training strategy that effectively integrates multi-task learning. The first stage is to establish a robust geometric foundation for our 3D encoder, and we train model only on the \textit{Task 1} (Part layout prediction). In addition, we initialize the point encoder with weights from P3SAM~\cite{ma2025p3}, a model pre-trained on large-scale part segmentation. This stage provides our encoder with a strong prior for identifying part-level geometry. In the second stage, we initialize the point encoder and projector with the weights from the first stage. We then perform SFT on model using all three tasks. This refines the model's understanding by training it to focus on kinematic reasoning.

This multi-task, multi-stage training strategy is crucial as we verify in ablation studies (Sec.~\ref{sec:exp_ablation}). By first grounding the 3D encoder on part-centric feature learning, we establish a high-quality weight initialization that significantly stabilizes the subsequent, more complex multi-task SFT. It effectively decouples geometric understanding from kinematic reasoning during initial training, enabling the model to learn their intricate relationship more robustly.

\subsection{Training Corpus for \methodname}
\label{sec:ArtLLMDataset}



To training \methodname, we construct a new large-scale dataset by aggregating and refining existing articulation datasets as well as procedurally generated data. Our dataset comprises objects from established benchmarks: PartNet-Mobility~\cite{xiang2020sapien} and PhysX3D~\cite{cao2025physx}. To enhance scale and diversity, we supplement these with 12k synthetic assets generated via the procedural method of Infinite-Mobility~\cite{lian2025infinite}.

\begin{table}[t]
\centering
\caption{Statistics of our curated dataset for \methodname training. }
\label{tab:dataset_statistics}
\begin{tabular}{lcc}
    \toprule
    \textbf{Dataset Source} & \textbf{\#Objects} & \textbf{\#Categories} \\
    \midrule
    PartNet-Mobility~\cite{xiang2020sapien} & 2168 & 43 \\
    PhysX3D~\cite{cao2025physx} & 7672 & 23 \\
    Infinite-Mobility~\cite{lian2025infinite} & 10833 & 13 \\
    \midrule
    \textbf{Total} & \textbf{20673} & \textbf{43} \\
    \bottomrule
\end{tabular}
\vspace{-1.5em}
\end{table}

We then perform data preprocessing on the collected raw articulation assets, including:
\begin{itemize}
    \item \textit{Filtering}: We remove objects with more than 20 joints and exclude categories containing excessively small parts (e.g., keyboard, remote). Small components, such as buttons, are also filtered according to part volume thresholds.
    \item \textit{Structure Simplification}: All fixed joints are removed, and their connected links are merged. For screw joints, which are usually represented as a combination of revolute and prismatic joints in URDF files, we merge them into a single screw joint to reduce prediction complexity.
    \item \textit{Normalization}: All joint parameters are transformed into the global coordinate frame, and normalized to the range $[-0.9,0.9]$ together with geometry.
    \item \textit{Normal Correction}: For models in PartNet-Mobility~\cite{xiang2020sapien} with incorrect surface normals, we apply watertight reconstruction to obtain accurate surface normals.
\end{itemize}

This pipeline yields a dataset of \textbf{20,673 articulated objects across 43 categories}. Table~\ref{tab:dataset_statistics} shows its statistics.



\subsection{Part-Aware Geometry Synthesis}
\label{sec:part_gen}

Our framework generates a structural blueprint—a layout of part bounding boxes—that can be seamlessly integrated with recent part-based generative models. Methods like OmniPart~\cite{yang2025omnipart} and XPart~\cite{yan2025x} are particularly well-suited, as they can condition geometry synthesis on bounding box inputs. For this work, we adopt XPart as our geometry generation backbone.

However, when predicted bounding boxes do not perfectly encompass the ground-truth part geometry, generated parts might be truncated or incomplete. To mitigate this, we introduce a robust \textbf{bounding box expansion} step that ensures complete geometric coverage. First, we iterate through every point in the input object's point cloud. Any point not contained within any predicted bounding box is assigned to its nearest box based on Euclidean distance. Subsequently, each bounding box is expanded just enough to tightly enclose all points newly assigned to it. This simple yet effective mechanism guarantees that the entire object point cloud is covered, preventing geometric artifacts and ensuring the fidelity of the final generated parts.

\begin{figure}[t]
    \centering
        \includegraphics[width=1.0\linewidth]{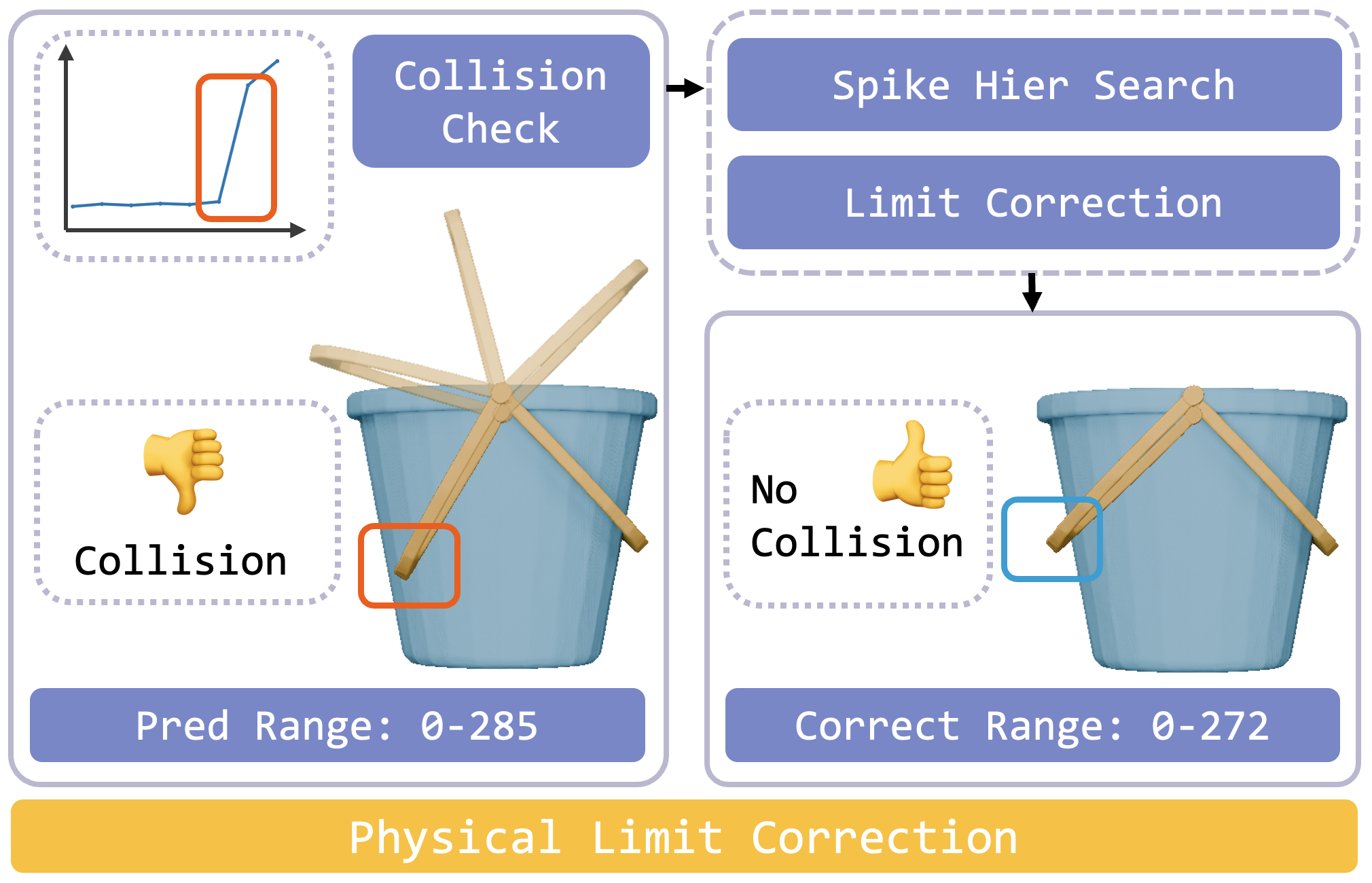}
    \caption{\textbf{Physical limit calcualtion.} Illustration for our physical based limit correction process.}
    \label{fig:physical_limit_calculation}
    \vspace{-1.5em}
\end{figure}


\begin{figure*}[t]
    \centering
        \includegraphics[width=\textwidth]{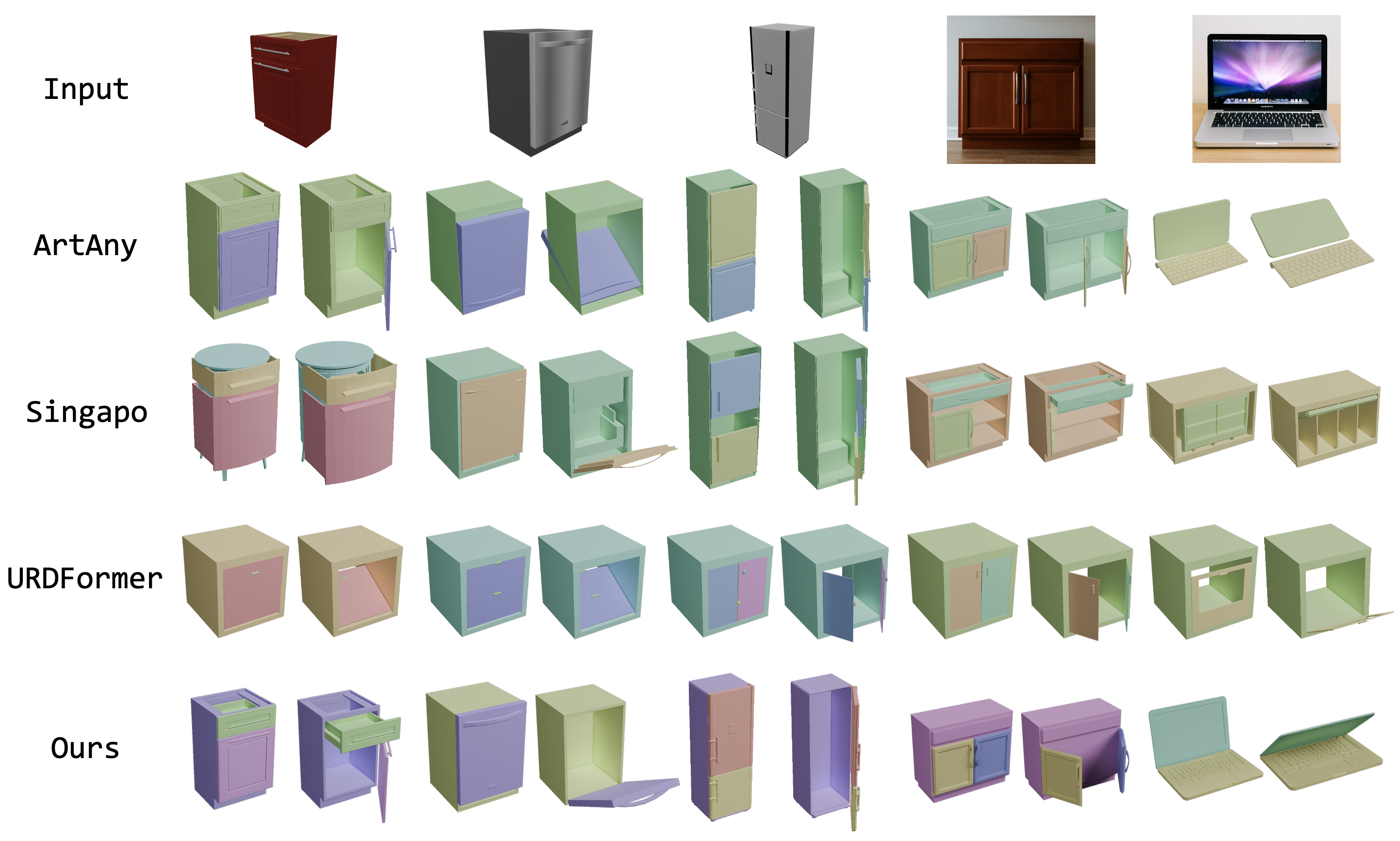}
    \caption{\textbf{Qualitative Comparison}. Baseline methods rely on retrieving parts from a fixed asset library, hence often fail to recover accurate geometry and frequently generate incorrect articulations with mismatched joint types or misaligned joint positions. In contrast, our approach produces geometry that closely matches the input and recovers correct, coherent articulations.}
    \label{fig:exp_comparison}
    \vspace{-1.5em}
\end{figure*}


\subsection{Physically-Constrained Joint Limit Correction}
\label{sec:limit_correction}

When predicting joint limits, the model relies solely on the geometric state at a single timestep, which limits its ability to perceive dynamic motion. 
This can lead to inter-part collisions during articulation, thereby compromising physical realism.
To address this issue, we introduce a post-processing correction step that refines joint limits based on collision detection.

Our method is illustrated in Figure~\ref{fig:physical_limit_calculation}. For a given revolute joint, we articulate its child part through its initially predicted range and compute the collision volume against all other static parts at discrete steps. Significant collisions manifest as sharp spikes in the derivative of this collision volume with respect to the joint angle. We first identify a coarse angular window containing a spike and then perform a hierarchical search within this window to pinpoint the precise angle of initial contact. This angle is then set as the refined, collision-free joint limit. A similar procedure is applied to prismatic joints based on translational distance.

By incorporating this physics-based refinement, we effectively prevent self-collisions. This enhances the stability and realism of the asset in physical simulations, making it more reliable for downstream applications like robotic manipulation.

\section{Experiments}%
\label{sec:exp}


In this section, we compare our approach with the current state-of-the-art articulation object generation methods to demonstrate its superior performance. We also conduct ablation studies on the designed modules to validate the soundness of our design. Finally, we further verify the practicality of our method in real robotic scenarios. 

\subsection{Evaluation Dataset}
We adopt the PartNet-Mobility~\cite{xiang2020sapien} dataset for evaluation. Following the data split proposed in SINGAPO~\cite{liu2024singapo}, we select 7 categories (Storage, Table, Refrigerator, Dishwasher, Oven, Washer, and Microwave), comprising a total of 77 objects, as our test set. These objects are held out during training. In addition, we adopt several real-world images, including objects outside these seven categories, to further assess the model’s generalization ability.

\begin{table*}[htbp]
  \centering
  \caption{\textbf{Quantitative Comparison.}
    We evaluate all methods on the seven categories of the PartNet-Mobility dataset (splitted by SINGAPO~\cite{liu2024singapo}). Metrics are computed per category, and the final score is obtained by an average over all seven categories. * denotes retraining on our dataset. 
    Our method attains high performance, highlighting its strong ability to recover accurate articulated structures. 
    }
  \label{tab:exp_quantitative_comparison}
    \begin{tabular}{c|ccccccccc}
    \toprule
    Method & mIoU & CD & Type Acc & Joint-Axis-Err & Joint-Pivot-Err & Joint-Range-IoU & Graph Acc & Time(s) \\
    \midrule
    \multirow{1}[0]{*}{Ours}
        & \textbf{0.6884} & \textbf{0.028} & \textbf{0.9084} & \textbf{0.1271} & \textbf{0.0801} & 0.7398 & \textbf{0.7741} & \textbf{19} \\
    \midrule
    \multirow{1}[0]{*}{ArtAny} 
        & 0.3381 & 0.072 & 0.8457 & 0.4529 & 0.5361 & \textbf{0.8653} & 0.6142 & 522 \\
    \midrule
    \multirow{1}[0]{*}{Singapo} 
        & 0.4330 & 0.044 & 0.7649 & 0.2445 & 0.2567 & 0.5256 & 0.4564 & 84 \\
     \midrule
     \multirow{1}[0]{*}{Singapo*} 
        & 0.4705 & 0.048 & 0.9065 & 0.2463 & 0.1465 & 0.6184 & 0.6851 & 84 \\
     \midrule
     \multirow{1}[0]{*}{urdformer} 
        & 0.1225 & 0.249 & 0.6068 & 0.7377 & 0.6095 & 0.7032 & 0.0791 & 183 \\
    \bottomrule
    \end{tabular}
    \vspace{-1.5em}
\end{table*}


\subsection{Evaluation Metrics}
For the model’s predicted results and the ground-truth data, we first perform scale and offset alignment of the objects, along with coordinate conversion (e.g., from z-up to y-up). Next, similar to ~\cite{liu2024singapo}, we apply the Hungarian matching algorithm to match parts based on the distances between their centers. Finally, using the established part correspondences, we can derive the matching relationships for the joints.

To thoroughly evaluate the model’s capability, we adopt multiple metrics for part prediction and articulation prediction. For part layout, we compute the mIoU between parts.
For articulation prediction, we first compute the joint type accuracy, then follow FreeArt3D~\cite{chen2025freeart3d} to adopt angle and minimum distance betwe axis to evaluate the axis prediction. Finally, we evaluate the accuracy of limit prediction by computing the IoU of the predicted limit ranges.
We also evaluate the kinematic hierarchy by constructing a directed graph from links and joints, and computing graph accuracy. 

\subsection{Comparisons}

\paragraph{Comparison Methods}
We conduct a comprehensive comparison between our method and recent state-of-the-art baselines, including URDFormer~\cite{chen2024urdformer}, Singapo~\cite{liu2024singapo}, and Articulate-Anything~\cite{le2024articulate}.
Since URDFormer is trained only on five categories, we restrict our evaluation to these categories and explicitly denote this setting in the table. Besides, both Singapo and URDFormer treat handles as separate parts and connect them to parent components using fixed joints. For a consistent evaluation protocol, we remove these fixed joints and merge the corresponding child and parent links. For Articulate-Anything, we evaluate the model with the GPT-4o API. To ensure fairness, we also remove the ground-truth object parts from the retrieval library to ensure fairness. As for our method, we adopt Hunyuan3D 3.0 to generate accurate geometry from the input image, and sample surface points as the input of our pipeline.

\noindent\textbf{Results} As shown in Fig~\ref{fig:exp_comparison}, our method is able to generate articulated objects whose shapes closely match the input images. In contrast, URDFormer~\cite{chen2024urdformer} relies on a fixed assumption that objects consist of an external frame and internal components, which leads to results that differ significantly from the ground truth appearance. Moreover, it fails to accurately predict the number of articulated parts and their rotation directions, resulting in very limited articulation structures. Although Articulate Anything~\cite{le2024articulate} and SINGAPO~\cite{liu2024singapo} can retrieve nearly identical geometry from their part databases, both methods exhibit clear limitations in joint prediction: Articulate Anything often misidentifies the axis direction, producing incorrect motion types, while SINGAPO frequently predicts inaccurate part scales and axis positions, causing the retrieved parts to be misaligned or poorly sized.
On real-world examples, URDFormer still produces highly limited results. Articulate Anything can reconstruct almost identical geometry, but continues to suffer from substantial errors in axis localization. SINGAPO fails to retrieve accurate geometry and often misses essential parts.

Quantitative results in Table ~\ref{tab:exp_quantitative_comparison} show that our method achieves a clear advantage over existing approaches in part layout prediction, joint accuracy, and hierarchical structure modeling. Although Articulate Anything performs well with rule-based limit prediction, it suffers from large axis-position errors. SINGAPO achieves reasonable performance on limit prediction but is fundamentally constrained by its small training category set. After retraining the SINGAPO model, all metrics showed significant improvement, but a noticeable gap remains compared to our method, demonstrating the effectiveness of our model architecture.
Furthermore, we report the inference time of each method, demonstrating that our method is also significantly faster, providing an efficient and scalable pipeline for generating articulated assets for large-scale simulation environments.

\begin{table}[tbp]
  \centering
  \caption{\textbf{Quantitative Ablation. } Ablation experiments evaluating the impact of our key components. }
  \label{tab:exp_quantitative_ablation}
    \begin{tabular}{c|cccccccc}
    \toprule
     & IoU & TA & JAE  & JPE & JRI & GA \\
    \midrule
    \multirow{1}[0]{*}{Full}
        & \textbf{0.473} & \textbf{0.898} & \textbf{0.141} & 0.135 & \textbf{0.582} & \textbf{0.780} \\
    \midrule
    \multirow{1}[0]{*}{A} 
        & 0.352 & 0.823 & 0.277 & 0.235 & 0.575 & 0.775 \\
    \midrule
    \multirow{1}[0]{*}{B} 
        & 0.464 & 0.825 & 0.289 & \textbf{0.131} & 0.510 & 0.737 \\
    \midrule
    \multirow{1}[0]{*}{C} 
        & 0.412 & 0.894 & 0.142 & 0.138 & 0.577 & 0.754 \\
    \midrule
    \multirow{1}[0]{*}{D} 
        & 0.463 & 0.890 & 0.143 & 0.175 & 0.511 & 0.780 \\
    \bottomrule
    \end{tabular}
    \vspace{-1.5em}
\end{table}


\subsection{Ablation}
\label{sec:exp_ablation}


We further conducted ablation studies to demonstrate the effectiveness of our proposed components. Due to the large computation cost, we train the model only on the PartNet-Mobility~\cite{xiang2020sapien} training set for 30 epochs for each ablation variant. To more comprehensively evaluate the model’s capability, we additionally select 2 objects in each category of PartNet-Mobility based on the split of SINGAPO~\cite{liu2024singapo}, totaling 144 objects. 

In Experiment A, directly predicting continuous part layouts and articulation parameters significantly weakens the model’s ability to infer coordinate and direction-related attributes, highlighting the difficulty of auto-regressive continuous prediction. In Experiment B, removing the multi-task setup slightly improves axis-direction prediction but degrades all other metrics, indicating that multi-task learning with varied difficulty reinforces part and articulation understanding. In Experiment C, eliminating random scaling and rotation leads to lower part-IoU performance, showing that 3D augmentation enhances spatial perception of part position and scale. In Experiment D, removing the multi-stage training and using P3SAM~\cite{ma2025p3} initialized point encoders lowers both part and joint prediction accuracy, demonstrating that pretraining on part-layout prediction yields superior encoder initialization.

\begin{figure}[t]
    \centering
        \includegraphics[width=1.0\linewidth]{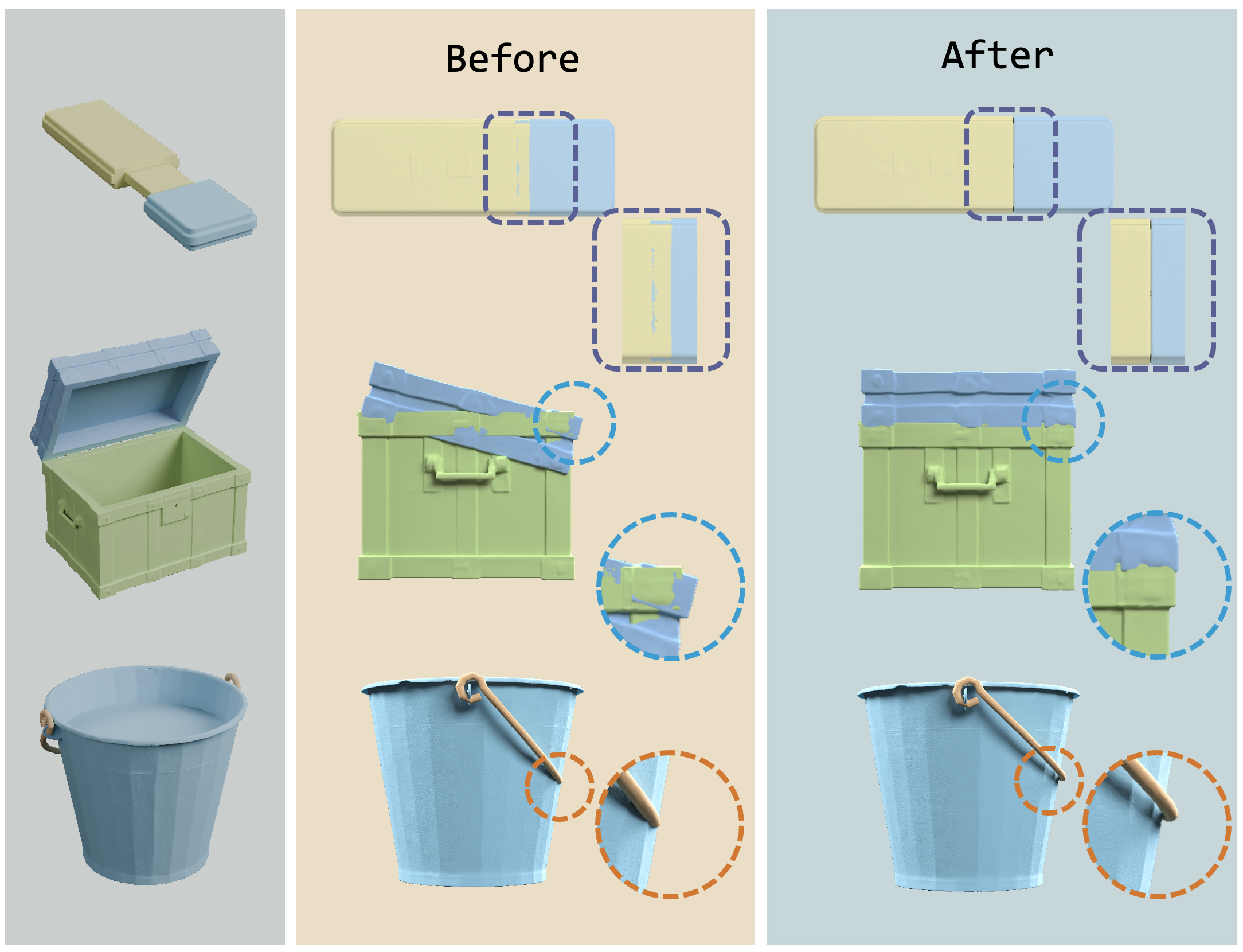}
    \caption{Qualitative result for physical limit correction. Before correction, the predicted joint ranges cause noticeable self-collisions during articulation. After applying our physics-based limit refinement, the articulated parts move smoothly without collision, yielding physically plausible and stable motion.}
    \label{fig:exp_physical_qualitative}
    \vspace{-1.5em}
\end{figure}

For the physical constraint–based limit correction proposed in the part geometry generation stage, we show qualitative results in Figure~\ref {fig:exp_physical_qualitative}. In several cases where predicted limits cause self-collision, our correction strategy effectively adjusts the limits to ensure that the articulated parts remain collision-free. This further improves the stability and realism of the generated assets when used for training in simulation environments.

\subsection{Application in Robotic Area}

To further demonstrate the value of our proposed method for generating articulated object assets in robotics learning, we conducted a real2sim evaluation. We first teleoperated a Franka Panda robot arm equipped with a Robotiq gripper in the real world to perform several tasks, and recorded the full pose sequence of the execution. Next, we used our \methodname pipeline to reconstruct the real objects as articulated assets suitable for simulation, placed them into a simulated scene, and replayed the recorded pose sequence with the simulated robot. By assessing whether the simulated objects exhibit the same articulation behavior as in the real environment, we evaluate how faithfully our generated assets preserve real-world articulation properties.

\begin{figure}[t]
    \centering
        \includegraphics[width=\linewidth]{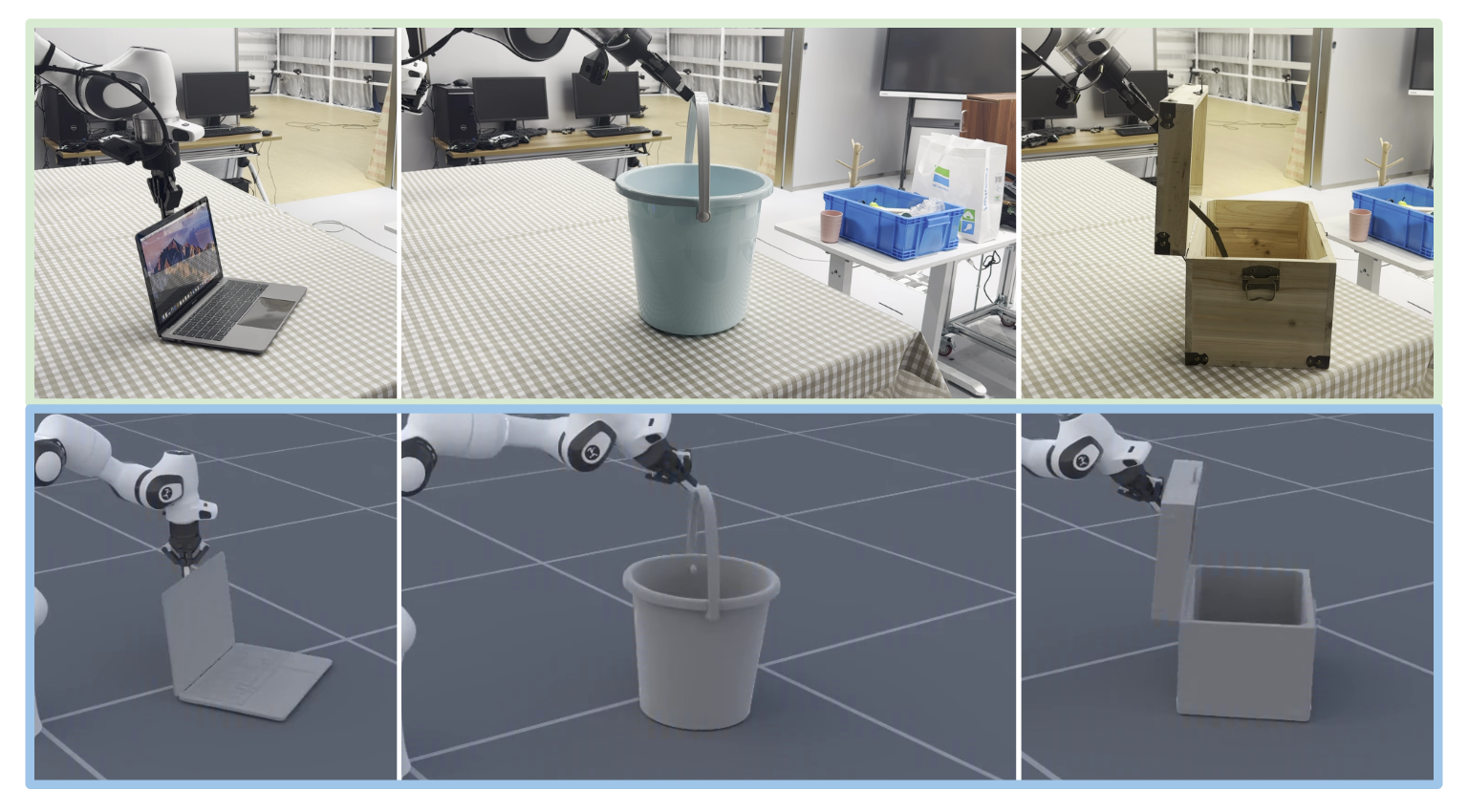}
    \caption{We teleoperate a Franka Panda robot to execute three articulation tasks and record its pose trajectories. Using \methodname, we reconstruct the corresponding articulated assets from real scenes and replay the trajectories in simulation. The simulated objects reproduce the real articulation behavior, showing that our generated assets accurately capture real-world kinematics and joint constraints.}
    \label{fig:application}
    \vspace{-1.5em}
\end{figure}

We tested three tasks: closing a laptop, closing a box, and moving a bucket handle. We employed Hunyuan3D 3.0 to reconstruct accurate 3D object geometry from video frames, and used our pipeline to generate URDF-format articulated assets. In SAPIEN~\cite{xiang2020sapien}, we arranged the generated assets and the robot in appropriate positions and executed the replay. As shown in Fig~\ref{fig:application}, all three tasks were successfully reproduced in the simulator, demonstrating the high fidelity of our articulated asset generation in terms of structure, joint constraints, and resulting motion behavior.

\section{Limitation}%
\label{sec:limitation}


Although our method efficiently generates high-quality articulated object assets, it has several limitations.
First, though training on a large curated dataset, the diversity of object categories is still limited. As a result, the model generalizes well to common household items but struggles with more complex categories such as vehicles or robots. Future work could incorporate open-vocabulary approaches like Kinematify~\cite{wang2025kinematify} to expand category coverage and enable broader object modeling.
Second, the framework does not jointly model physical properties. Training on large-scale datasets with annotated physical attributes could equip the model with physics-aware prediction capabilities, which we leave for future work.
\section{Conclusion}%
\label{sec:conclusion}



In this work, we present \methodname, an efficient framework capable of rapidly generating articulated objects from various modalities such as a single image or text description. Our approach models part layouts and articulation parameters in an autoregressive, language-based manner, enabling flexible representation of objects with varying numbers of joints and topologies. The proposed data discretization design effectively improves numerical stability in next-token prediction. Furthermore, \methodname integrates seamlessly with existing high-quality part generation modules, allowing us to synthesize geometrically detailed and diverse parts, thus overcoming the low-novel geometry limitation common in retrieval-based methods. In addition, our physical constraint-based limit correction mechanism effectively mitigates mesh collisions, producing physically grounded assets that are valuable for robotic simulation and related downstream tasks. Overall, \methodname provides a stable and generalizable pipeline for articulated object generation, with strong potential to narrow the real-to-simulation gap, accelerate the creation of high-fidelity digital twins, thus enable scalable robot learning.
\section{Acknowledgement}
\label{sec:ack}

This work was supported by the Shanghai Pujiang Program (24PJA080), the MoE Key Lab of Intelligent Perception and
Human-Machine Collaboration (ShanghaiTech University), the Shanghai Frontiers Science Center of
Human-centered Artificial Intelligence, and the HPC Platform of ShanghaiTech University.
We gratefully acknowledge the invaluable discussion and feedback provided by \textbf{Chunshi Wang}, \textbf{Junliang Ye}, \textbf{Yunhan Yang} from the Tencent Hunyuan3D Team, \textbf{Xinyu Lian} from the Shanghai AI Lab, and \textbf{Kaixin Yao}, \textbf{Zhehao Shen} from ShanghaiTech University.
{
    \small
    \bibliographystyle{ieeenat_fullname}
    \bibliography{main}
}

\clearpage
\maketitlesupplementary


\section{Implementation Details}

Our \methodname training is built upon the codebases of LLaMA-Factory~\cite{zheng2024llamafactory} and SpatialLM~\cite{mao2025spatiallm}. In each stage, the cross-entropy loss is used as the optimization objective for SFT. The multi-task data mixing ratio is set to 3:2:5. We employ a cosine learning rate scheduler with a maximum rate of 1e-5 and a warmup ratio of 0.03. 

During training, we augment 3D input point clouds with random scaling and rotations, applied to each sample with a probability of 0.75. For random scaling, we sample a scale factor $s \in [0.8, 1.05]$, and apply it to the object. For random rotation, we randomly select a rotation angle $\theta$ from $90, 180, 270$, and rotate the object along the y axis. These transformations will also be applied on the part layout and articulations. 

In Stage 1, we train on mixed data using Task 1 with 8 H20 GPUs for 50 epochs to obtain initial weights for the point encoder and projector, which takes approximately 8 hours. In Stage 2, we train the model on mixed data using all three tasks with 8 H20 GPUs for 30 epochs, requiring 15 hours. 

The part generation model uses the publicly released version of XPart~\cite{yan2025x}. After generating the link geometry, we combine it with the previously predicted articulations and export the result in URDF format, enabling seamless integration into simulators for further analysis and simulation.

\section{ArtLLM Template}

Here we present the text prompts used for training ArtLLM in our multi-task training. Please note that our ArtLLM is targeted at specific tasks. Therefore, we keep the text prompts as concise as possible to reduce token usage while still enabling clear differentiation between multi-task.
\begin{verbatim}
# Task 1: Part Layout Prediction
Detect part boxes.

# Task 2: Kinematic Prediction
Given part boxes, detect joints.

# Task 3: End-to-End Prediction
Detect part boxes and joints.
\end{verbatim}
Next, we provide the output format template for ArtLLM. Following SpatialLM~\cite{mao2025spatiallm}, we adopt a concise yet sufficiently informative template as the LLM's output format, which includes the part layout and the four types of joint articulations including RevoluteJoint, CountinuousJoint, ScrewJoint, and Prismatic Joint. 

Part bounding box:
\begin{verbatim}
@dataclass
class BBox:
    min_x: int
    min_y: int
    min_z: int
    max_x: int
    max_y: int
    max_z: int
\end{verbatim}

Revolute joint:
\begin{verbatim}
@dataclass
class RevoluteJoint:
    parent_box_id: int
    child_box_id: int
    axis_direction: int
    axis_position: [int, int, int]
    rotation_limit: [int, int]
\end{verbatim}

Continuous joint:
\begin{verbatim}
@dataclass
class ContinuousJoint:
    parent_box_id: int
    child_box_id: int
    axis_direction: int
    axis_position: [int, int, int]
\end{verbatim}

Screw joint:
\begin{verbatim}
@dataclass
class ScrewJoint:
    parent_box_id: int
    child_box_id: int
    axis_direction: int
    axis_position: [int, int, int]
    translation_limit: [int, int]
\end{verbatim}

Prismatic joint:
\begin{verbatim}
@dataclass
class PrismaticJoint:
    parent_box_id: int
    child_box_id: int
    axis_direction: int
    translation_limit: [int, int]
\end{verbatim}

\section{Detail of Training Dataset}
In this section, we further provide the detailed information of our curated dataset used for training \methodname. 

\begin{figure*}[t]
    \centering
        \includegraphics[width=\textwidth]{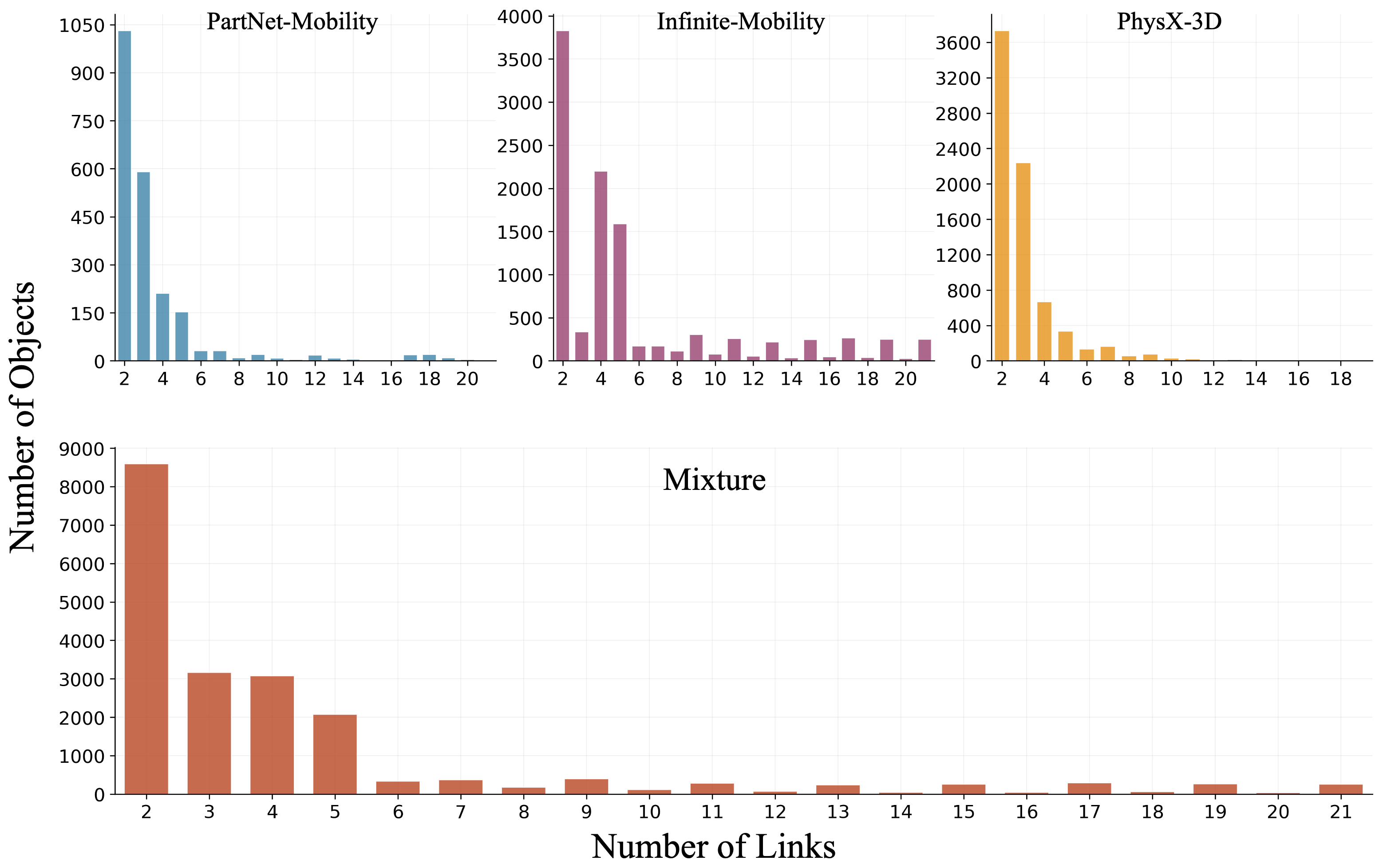}
    \caption{Part number statistic of our curated dataset for training \textbf{\methodname}}
    \label{fig:part_number_statistic}
    \vspace{-1.5em}
\end{figure*}

\subsection{Detail Statistic}
We begin by providing a more detailed statistical analysis of the dataset. 

As shown in Figure~\ref{fig:part_number_statistic}, we first statistic the distribution of samples by the number of links. The curated dataset filtered sample with more than 20 joints, so the links are no more than 21. Although most objects in the dataset contain a relatively small number of parts, we have collected a sufficient number of samples with larger part counts. This ensures that our model can effectively handle objects with varying numbers of parts and generalize well across different levels of structural complexity.

\subsection{Layout and articulation sort}
Since the ordering of data is crucial for the training stability of autoregressive models such as LLMs, we further apply sorting operations to both the layout and articulation data to convert unordered bounding boxes and joints into ordered sequences.
For the layout, following the approach used in OmniPart~\cite{yang2025omnipart}, we sort the quantized bounding boxes based on their minimum coordinates in z–y–x order.
For articulation, we sort all joints according to the IDs of their child links in ascending order.

By incorporating this sorting procedure, we ensure that each sample is represented as a deterministically ordered sequence, eliminating ambiguity in ordering and improving the stability of LLM training and inference.

\subsection{Example Training Data}
To facilitate a clearer understanding of the input and output formats of \methodname, we present a randomly selected sample from our training dataset, which is in ShareGPT format. The example includes the raw data sample with the continuous value, and the data sample with the discretized value that is actually used during training.

Raw data sample with continuous value: 
\begin{verbatim}
{"conversations": [
  {
    "from": "human",
    "value": 
    "<point_cloud>
    Detect part boxes and joints. 
    The reference code is as followed: 
    @dataclass
    class BBox:    
        min_x: int    
        min_y: int    
        min_z: int    
        max_x: int    
        max_y: int    
        max_z: int
        
    @dataclass
    class RevoluteJoint:    
        parent_box_id: int    
        child_box_id: int    
        axis_direction: int    
        axis_position: [int, int, int]    
        rotation_limit: [int, int]
        
    @dataclass
    class ContinuousJoint:    
        parent_box_id: int    
        child_box_id: int    
        axis_direction: int    
        axis_position: [int, int, int]
        
    @dataclass
    class ScrewJoint:    
        parent_box_id: int    
        child_box_id: int    
        axis_direction: int    
        axis_position: [int, int, int]    
        translation_limit: [int, int]
    
    @dataclass
    class PrismaticJoint:    
        parent_box_id: int    
        child_box_id: int    
        axis_direction: int    
        translation_limit: [int, int]"
  },
  {
    "from": "gpt",
    "value": "<|layout_s|>
    bbox_0=BBox(
        0.3900, -0.3525, 0.2381,
        0.8700, 0.2925, 0.3019
    )
    bbox_1=BBox(
        -0.0000, -0.3525, 0.2381,
        0.3300, 0.2925, 0.3019
    )
    bbox_2=BBox(
        -0.3300, -0.3525, 0.2381,
        -0.0000, 0.2925, 0.3019
    )
    bbox_3=BBox(
        -0.8700, -0.3525, 0.2381,
        -0.3900, 0.2925, 0.3019
    )
    bbox_4=BBox(
        0.3900, 0.3075, -0.1819,
        0.8700, 0.4875, 0.2953
    )
    bbox_5=BBox(
        -0.8700, 0.3075, -0.1819,
        -0.3900, 0.4875, 0.2953
    )
    bbox_6=BBox(
        -0.9000, -0.5175, -0.3019,
        0.9000, 0.5175, 0.2606
    )  
    <|layout_e|>
    <|art_s|>
    joint_0=RevoluteJoint(
        6, 0, [0, 1, 0],
        [0.8672, -0.0300, 0.2400],
        [0, 90]
    )
    joint_1=RevoluteJoint(
        6, 1, [0, 1, 0],
        [0.3280, -0.0300, 0.2396],
        [0, 90]
    )
    joint_2=RevoluteJoint(
        6, 2, [0, 1, 0],
        [-0.3283, -0.0300, 0.2384],
        [0, -90]
    )
    joint_3=RevoluteJoint(
        6, 3, [0, 1, 0],
        [-0.8688, -0.0300, 0.2394],
        [0, -90]
    )
    joint_4=PrismaticJoint(
        6, 4, [0, 0, 1],
        [0.0000, 0.3717]
    )
    joint_5=PrismaticJoint(
        6, 5, [0, 0, 1],
        [0.0000, 0.3717]
    )
    <|art_e|>"
  }
],
"point_clouds": ["46145/pcd/46145.ply"]
}
\end{verbatim}

The discretized value actually used in our training. We use special tokens to represent the bounding box coordinate, axis direction, axis origin, and axis limit range. 

\begin{verbatim}
<|layout_start|>
    bbox_0 = BBox(
        <P_6>, <P_30>, <P_44>, 
        <P_122>, <P_98>, <P_81>
    )
    bbox_1 = BBox(
        <P_8>, <P_83>, <P_52>, 
        <P_40>, <P_96>, <P_83>
    )
    bbox_2 = BBox(
        <P_88>, <P_83>, <P_52>, 
        <P_120>, <P_96>, <P_83>
    )
    bbox_3 = BBox(
        <P_8>, <P_41>, <P_79>, 
        <P_40>, <P_83>, <P_84>
    )
    bbox_4 = BBox(
        <P_42>, <P_41>, <P_79>, 
        <P_64>, <P_83>, <P_84>
    )
    bbox_5 = BBox(
        <P_63>, <P_41>, <P_79>, 
        <P_86>, <P_83>, <P_84>
    )
    bbox_6 = BBox(
        <P_88>, <P_41>, <P_79>, 
        <P_120>, <P_83>, <P_84>
    )
<|layout_end|>
<|art_start|>
    joint_0 = PrismaticJoint(
        0, 1, <D_4>, 
        [<LT_32>, <LT_38>]
    )
    joint_1 = PrismaticJoint(
        0, 2, <D_4>, 
        [<LT_32>, <LT_38>]
    )
    joint_2 = RevoluteJoint(
        0, 3, <D_2>, 
        [<P_8>, <P_62>, <P_79>], 
        [<LR_18>, <LR_24>]
    )
    joint_3 = RevoluteJoint(
        0, 4, <D_2>, 
        [<P_42>, <P_62>, <P_79>], 
        [<LR_18>, <LR_24>]
    )
    joint_4 = RevoluteJoint(
        0, 5, <D_2>, 
        [<P_84>, <P_62>, <P_79>], 
        [<LR_24>, <LR_30>]
    )
    joint_5 = RevoluteJoint(
        0, 6, <D_2>, 
        [<P_119>, <P_62>, <P_79>], 
        [<LR_24>, <LR_30>]
    )
<|art_end|>
\end{verbatim}

During training, the point cloud is replaced to point tokens produced by the point cloud encoder. 

\begin{table*}[htbp]
  \centering
  \caption{\textbf{Quantitative Comparison.}
    We show the metrics on each category of the 3 method against our method. The '-' denotes that no limit is predicted due to the error joint type. 
  }
  \label{tab:exp_quantitative_comparison_category}
  \begin{tabular}{c|c|ccccccccc}
    \toprule
    Category & Method & mIoU & Type Acc & Joint-Axis-Err & Joint-Pivot-Err & Range-IoU & Graph Acc \\
    \midrule
    
    \multirow{4}{*}{Table}
        & Ours
        & 0.4528 & 0.7000 & 0.1125 & \textbf{0.1393} & 0.6451 & \textbf{0.7000} \\
        & ArtAnything
        & 0.2803 & 0.7500 & 0.2992 & 0.1458 & \textbf{0.8029} & 0.2500 \\
        & Singapo
        & \textbf{0.4716} & \textbf{1.0000} & \textbf{0.0017} & 0.1992 & 0.4906 & 0.4000 \\
        & URDFormer
        & 0.0999 & 0.6000 & 1.0472 & 0.8528 & 0.5941 & 0.1000 \\
    \midrule

    \multirow{4}{*}{StorageFurniture}
        & Ours
        & \textbf{0.7759} & 0.9091 & \textbf{0.1771} & \textbf{0.1192} & 0.7354 & \textbf{0.9091} \\
        & ArtAnything
        & 0.4464 & \textbf{0.9200} & 0.2537 & 0.4336 & \textbf{0.8555} & 0.8000 \\
        & Singapo
        & 0.4741 & 0.8545 & 0.2140 & 0.1760 & 0.6092 & 0.5455 \\
        & URDFormer
        & 0.0948 & 0.6481 & 0.6973 & 0.7580 & 0.8302 & 0.2037 \\
    \midrule

    \multirow{4}{*}{WashingMachine}
        & Ours
        & \textbf{0.7978} & \textbf{1.0000} & 0.0493 & \textbf{0.0281} & \textbf{0.9326} & \textbf{1.0000} \\
        & ArtAnything
        & 0.3889 & 1.0000 & \textbf{0.0000} & 1.0089 & 0.9191 & 0.0000 \\
        & Singapo
        & 0.3364 & 0.0000 & 1.2821 & 0.4992 & -- & 0.0000 \\
        & URDFormer
        & 0.0000 & 0.0000 & 0.9329 & 0.2462 & -- & 0.0000 \\
    \midrule

    \multirow{4}{*}{Dishwasher}
        & Ours
        & \textbf{0.6359} & \textbf{1.0000} & 0.2125 & \textbf{0.0965} & 0.8333 & \textbf{1.0000} \\
        & ArtAnything
        & 0.4761 & 1.0000 & 1.0472 & 0.0968 & \textbf{1.0000} & 1.0000 \\
        & Singapo
        & 0.4600 & 1.0000 & \textbf{0.0039} & 0.5999 & 0.6563 & 1.0000 \\
        & URDFormer
        & 0.1718 & 1.0000 & 0.5236 & 0.4210 & 0.9995 & 0.0000 \\
    \midrule

    \multirow{4}{*}{Refrigerator}
        & Ours
        & \textbf{0.5916} & 0.7500 & 0.1703 & 0.0948 & \textbf{1.0000} & \textbf{0.7500} \\
        & ArtAnything
        & 0.3391 & 0.7500 & 0.3927 & 0.2450 & 0.5000 & 0.7500 \\
        & Singapo
        & 0.2890 & \textbf{1.0000} & \textbf{0.0059} & \textbf{0.0301} & 0.5835 & 0.2500 \\
        & URDFormer
        & 0.1044 & 0.5000 & 0.7854 & 1.0901 & 0.4997 & 0.2500 \\
    \midrule

    \multirow{4}{*}{Microwave}
        & Ours
        & \textbf{0.9172} & \textbf{1.0000} & \textbf{0.0000} & \textbf{0.0710} & 0.5000 & \textbf{1.0000} \\
        & ArtAnything
        & 0.2327 & 1.0000 & 0.0000 & 1.4211 & \textbf{1.0000} & 1.0000 \\
        & Singapo
        & 0.5561 & 1.0000 & 0.0038 & 0.0718 & 0.6768 & 1.0000 \\
        & URDFormer
        & 0.1508 & 1.0000 & 0.0000 & 0.4621 & 0.9995 & 0.0000 \\
    \midrule

    \multirow{4}{*}{Oven}
        & Ours
        & \textbf{0.6474} & \textbf{1.0000} & 0.1682 & \textbf{0.0116} & 0.7721 & \textbf{1.0000} \\
        & ArtAnything
        & 0.2038 & 0.5000 & 1.1781 & 0.4022 & \textbf{0.9798} & 0.5000 \\
        & Singapo
        & 0.4441 & 0.5000 & \textbf{0.2002} & 0.2205 & 0.6628 & 0.0000 \\
        & URDFormer
        & 0.2363 & 0.5000 & 1.1781 & 0.4367 & 0.9995 & 0.0000 \\
    \bottomrule
  \end{tabular}
\end{table*}

\section{Detail of Experiments}

\subsection{Compute Resources}
In our comparison experiments, all methods were evaluated using the same computation resources. Specifically, model inference was conducted on an Ubuntu server equipped with 2 * Intel Xeon Silver 4210 processor and a single NVIDIA GeForce RTX 3090 GPU with 24GB VRAM. Our model training, including ablation experiments, are conducted on a server with 2 Intel Xeon Platinum 8476C processor, and 8 * NVIDIA H20 96GB GPU with NVLink.

\begin{figure*}[t]
    \centering
    \centering
        \includegraphics[width=\textwidth]{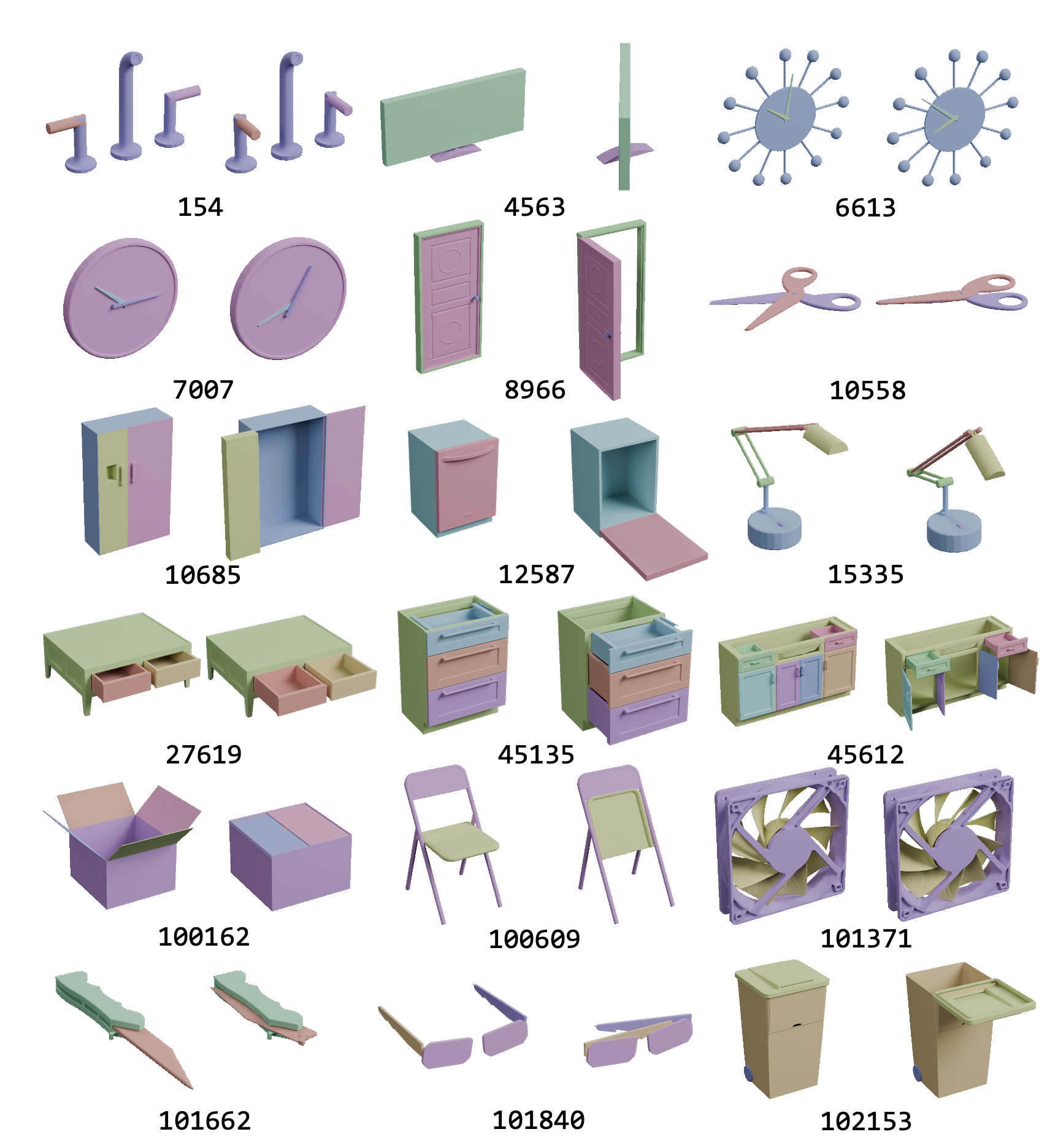}
    \caption{Additional results generated by our \textbf{\methodname}, from the test set of the PartNet-Mobility~\cite{xiang2020sapien} dataset. The input geometry is generated by Hunyuan3D 3.0 from rendered images. In each pair of images, the left one is the canoical state, and the right one is a sampled articulated state. }
    \label{fig:supp_gallery}
    \vspace{-1.5em}
\end{figure*}

\subsection{Metrics Compute}
Below we provide the detailed definitions of the evaluation metrics used in our experiments. After applying the scale and coordinate alignment, and use Hungarian matching algorithm to align parts and joints with ground-truth instances, we compute metrics for each component as follows:

For \textbf{Part Layout}, we use the mIoU as the evaluation metric, with 3d IoU defined as: 
\begin{equation}
    \mathrm{IoU} = \frac{V_{\text{inter}}}{V_{\text{union}}}
    = \frac{V_{\text{inter}}}{
        V_p + V_g - V_{\text{inter}}
    }, 
\end{equation}
and mIoU defined as 
\begin{equation}
    \mathrm{mIoU} = \frac{1}{N} \sum_{i=1}^{N} \mathrm{IoU}_i
\end{equation}

For \textbf{articulation}, the metrics including joint type accuracy, joint axis error, joint pivot error, and joint range IoU. For \textbf{joint type}, the evaluation metric is defined as: 
\begin{equation}
    acc_\text{type} = \begin{cases}
                    1, & \text{if } \text{type}(J_p) == \text{type}(J_g) \\
                    0, & \text{otherwise}
                    \end{cases}
\end{equation}
For \textbf{joint axis error}, we measure the angle between the predicted axis direction and the ground-truth direction. Since directions are equivalent up to sign, we compute the angle for both the original and reversed directions and take the smaller value as metric: 
\begin{equation}
    \begin{aligned}
e_{\text{axis}} = 
\min\Bigg(
    &\arccos\!\left( \frac{a_p \cdot a_g}{\lVert a_p\rVert_2 \lVert a_g\rVert_2} \right),\\
    &\arccos\!\left( \frac{-a_p \cdot a_g}{\lVert a_p\rVert_2 \lVert a_g\rVert_2} \right)
\Bigg),
\end{aligned}
\end{equation}
For \textbf{axis pivot error}, this evaluates the positional discrepancy between the predicted and ground-truth joint origin: 
\begin{equation}
    e_{\text{origin\_pos}}
    = 
    \frac{\lvert \mathbf{p} \cdot ( \mathbf{a}_p \times \mathbf{a}_g ) \rvert}
     {\lvert \mathbf{a}_p \times \mathbf{a}_g \rvert}.
\end{equation}
where $\mathbf{p} = x_p - x_g$. 

For \textbf{joint range IoU}, we similarly consider direction reversal. So we compute the error with respect to both the original and reversed directions and use the maximum one as the final metric:
\begin{equation}
    \mathrm{IoU}_{lim}=\max \Big(
        \mathrm{IoU}(r_p,\, r_g),\;
        \mathrm{IoU}(-r_p,\, r_g)
    \Big).
\end{equation}

For the hierarchical structure, we evaluate the \textbf{graph accuracy} by checking whether the predicted graph is isomorphic to the ground-truth graph, using lib networkx: 
\begin{equation}
    acc_\text{graph} = \begin{cases}
                    1, & \text{if } \text{graph}(p) == \text{graph}(g) \\
                    0, & \text{otherwise}
                    \end{cases}
\end{equation}

For all metrics, we first compute the metric for each individual object, then average across all objects within the same category, and finally average over all categories to obtain the final evaluation results.

\subsection{Comparison on each category}
To provide a clearer comparison across categories, we include a detailed table presenting the quantitative results for each category. As shown in Tab.~\ref{tab:exp_quantitative_comparison_category}, our method demonstrates clear superiority across most metrics and categories. Although Articulate Anything~\cite{le2024articulate} performs better on the limit range metric, it exhibits poor performance in part layout and joint origin prediction. SINGAPO~\cite{liu2024singapo}, on the other hand, occasionally shows advantages in joint axis prediction, but its performance is significantly weaker on the remaining metrics, particularly on joint origin.

\section{More Results}
Here we show the results generated by our \methodname on more categories in the PartNet-Mobility~\cite{xiang2020sapien} test set, the input point cloud for our model is still generated by Hunyuan3D 3.0. As shown in Fig.~\ref {fig:supp_gallery}, our model can generate realistic articulation assets from images across many categories with accurate geometry. 

\begin{figure}[t]
    \centering
    \centering
        \includegraphics[width=\linewidth]{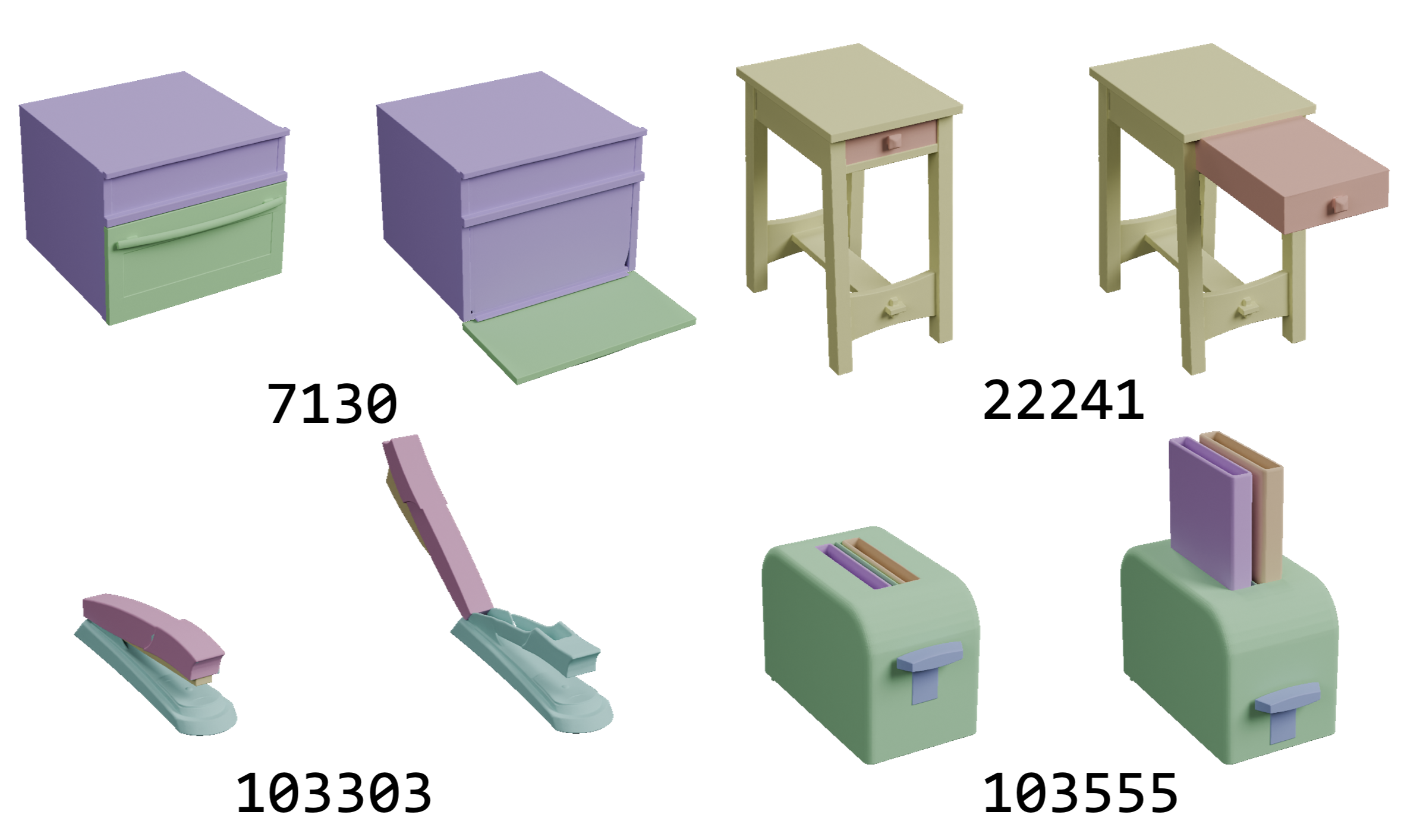}
    \caption{Failure cases analysis }
    \label{fig:failure_cases}
    \vspace{-1.5em}
\end{figure}

\section{Failure Cases Analysis}
Here, we also show some failure cases on the test set of PartNet-Mobility~\cite{xiang2020sapien} dataset. 

As shown in Fig.~\ref{fig:failure_cases}, in the case of 7130 and 22241, although our ArtLLM produces accurate layouts, XPart~\cite{yan2025x} fails to generate the internal concave structures when creating geometry, resulting in assets that lack realism.
In Case 103303, for two parts with a high degree of overlap, the generated geometries tend to intersect with each other.
In Case 103555, the detailed bread-rack structure is located inside the object and cannot be fully captured from a single image, causing it to be omitted during full-object generation. As a result, the articulation assets also fail to reproduce this internal fine structure, reducing realism.

To address these issues, future work may include fine-tuning XPart or training a 3D generative model capable of reconstructing internal object structures that are occluded in images.

\end{document}